\documentclass[letterpaper, 10 pt, conference]{ieeeconf}
\IEEEoverridecommandlockouts                              
\usepackage{times}
\usepackage{graphicx}
\usepackage[final]{changes}
\usepackage{xcolor}
\usepackage{amsmath}
\usepackage{amssymb}  
\usepackage{algorithm}
\usepackage[noend]{algorithmic}
\usepackage{cite}

\makeatletter
\newcommand\fs@spaceruled{\def\@fs@cfont{\bfseries}\let\@fs@capt\floatc@ruled
  \def\@fs@pre{\vspace{6pt}\hrule height.8pt depth0pt \kern2pt}%
  \def\@fs@post{\kern2pt\hrule\vspace{-10pt}\relax}%
  \def\@fs@mid{\kern2pt\hrule\kern2pt}%
  \let\@fs@iftopcapt\iftrue}
  \floatstyle{spaceruled}
  \restylefloat{algorithm}
\makeatother

\newcommand{\CodeComment}[1]{{\footnotesize{\emph{//#1}}}}
\newcommand{\Acronym}[1]{\ensuremath{{{\texttt{#1}}}}}
\newcommand{\Symbol}[1]{\ensuremath{\mathcal{#1}}}
\newcommand{\Function}[1]{\ensuremath{{\small{{\textsc{#1}}}}}}
\newcommand{\Constant}[1]{\ensuremath{{\texttt{#1}}}}
\newcommand{\Var}[1]{\ensuremath{{{\mathrm{#1}}}}}
\newcommand{\False}{\ensuremath{\Acronym{false}}}
\newcommand{\True}{\ensuremath{\Acronym{true}}}

\newcommand{\R}{\ensuremath{{\mathbb{R}}}}

\newcommand{\Subdiv}{\ensuremath{\Delta}}

\newcommand{\World}{\Symbol{W}}

\newcommand{\Goal}{\Symbol{G}}

\newcommand{\Traj}{\ensuremath{\zeta}}

\newcommand{\StateSpace}{\Symbol{S}}
\newcommand{\CtrlSpace}{\Symbol{U}}
\newcommand{\MotionEqs}{\ensuremath{f}}

\newcommand{\Sensor}{\Function{sense}}
\newcommand{\SensorGrid}{\ensuremath{\World_\Var{sensed}}}

\newcommand{\mtv}{\ensuremath{\eta}}

\newcommand{\Pair}[1]{\ensuremath{{\langle{#1} \rangle}}}

\newcommand{\Tree}{\Symbol{T}}

\newcommand{\Robot}{\Symbol{R}}

\newcommand{\Shape}{\ensuremath{\Symbol{P}}}
\newcommand{\Simulate}{\Function{simulate}}

\IEEEoverridecommandlockouts  
\title{Guided Sampling-Based Motion Planning with Dynamics in\\ Unknown Environments}

\author{Abhish Khanal \and Hoang-Dung Bui \and Gregory J. Stein \and Erion Plaku \thanks{The authors are with
    the Department of Computer Science, George Mason University, VA,
    USA.} }

\begin{document}
\maketitle

\begin{abstract}
Despite recent progress improving the efficiency and quality of motion planning, planning collision-free and dynamically-feasible trajectories in partially-mapped environments remains challenging, since constantly replanning as unseen obstacles are revealed during navigation both incurs significant computational expense and can introduce problematic oscillatory behavior. To improve the quality of motion planning in partial maps, this paper develops a framework that augments sampling-based motion planning to leverage a high-level discrete layer and prior solutions to guide motion-tree expansion during replanning, affording both (i) faster planning and (ii) improved solution coherence. Our framework shows significant improvements in runtime and solution distance when compared with other sampling-based motion planners.
 



  \end{abstract}

\section{Introduction}
\label{sec:Intro}

Motion planning is a foundational capability in robotics, with applications including transportation, self-driving, and search and rescue.
This task is made challenging by the presence of obstacles during deployment, which requires that the robot plan trajectories that avoid collision and pass through narrow passages to reach a faraway goal.
Moreover, motion planning must consider vehicle dynamics, so that the planned trajectories can be executed on a physical robot, whose physical limitations impose constraints on motion\replaced{.}{how the robot is able to move---e.g., restricting turning radius, direction of motion.}

The challenges of dynamically-feasible motion planning compound when the map is not known in advance.
So as to avoid the computational and practical challenges of envisioning what lies in unseen space, most algorithms for navigation and motion planning under uncertainty treat unseen space as unoccupied~\cite{koenig2002dstarlite,bohlin2000lazyprm,yang2022FARplanner}.
As the robot moves, it reveals obstacles via onboard sensors, adds them to its partial map of the environment, and replans upon discovering that its previously-planned trajectories may be infeasible.
Thus, the robot must continually replan as unseen space is revealed and so a practical motion planning system must be able to plan and replan quickly in the face of new information.

\begin{figure}
    \centering
    \includegraphics[width=0.8\columnwidth]{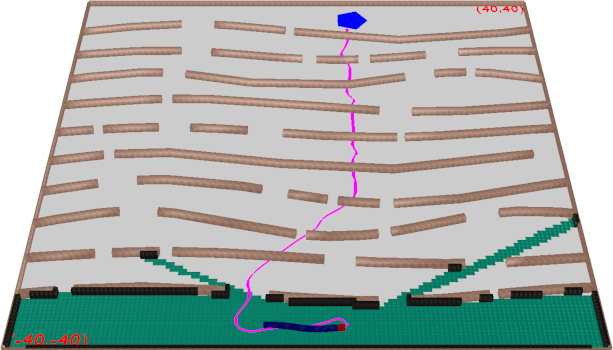}
    \caption{\textbf{A trajectory in a partially-mapped environment planned by our framework.} Our snake-like robot plans dynamically-feasible trajectories (magenta) that respect observed obstacles (black), moves in the observed free space (green),  and enters unseen space (tan) until reaching the goal. Our framework allows the robot to quickly replan as obstacles are revealed during navigation. Videos of solutions obtained by our framework on this and other scenes used in the experiments can be found at tinyurl.com/47ct55s6}
    \label{fig:MainScene}
    \vspace{-1em}
\end{figure}
To address these challenges, this work leverages sampling-based motion planning, in which collision-free, dynamically-feasible trajectories are extended to efficiently explore the state space until the goal is reached~\cite{book:MP,book:LaValle}.
While significant progress has been made in improving the speed and efficiency of sampling-based motion planning~\cite{RRT,KPIECE,GUST,PlakuClearance,MPDynamics2,ML1,NeuralRRTStar}, most of that progress has focused on fully-known environments, resulting in a slow replanning process and potentially poor closed-loop behavior.
In particular, owing to the randomness inherent in sampling-based planning, trajectories can non-trivially vary from small changes to the map or input, resulting in problematic oscillatory behavior as the robot iteratively navigates, reveals structure, and replans.

This work develops a new sampling-based framework particularly well-suited for motion planning with dynamics in unknown environments.
Our approach builds upon recent work using discrete to guide sample-based motion planning for fast, dynamically-feasible motion planning~\cite{GUST,PlakuClearance}, augmented so as to improve planning speed, solution quality, and robustness in partially-mapped environments.
Specifically, our planning framework leverages discrete search over an adaptive grid subdivision and prior solutions to guide the motion-tree exploration when invoked to plan from a new state, allowing for both (i) reduced replanning time and (ii) greater solution coherence to significantly reduce oscillatory behavior. In addition, our approach (iii) increases clearance from obstacles to improve robustness, and thereby improves the success rate in reaching the goal despite having only a partial map of the environment.
We demonstrate that our planning framework exhibits significant improvements in runtime and solution distance when compared with two other sampling-based motion planners: RRT~\cite{RRT} and GUST~\cite{GUST}.


\section{Related Work}

\paragraph{Motion Planning with Dynamics}
Motion planning has traditionally focused on fully-known environments. To account for the robot dynamics, sampling-based motion planning has often been used due to its computational efficiency. The idea is to selectively explore the vast motion space by constructing a motion tree whose branches correspond to collision-free and dynamically-feasible trajectories \cite{book:MP,book:LaValle}. \deleted{Over the years, numerous approaches have been proposed.} RRT~\cite{RRT,RRTRecent1} and its variants \cite{RRTreach,RRTtrans2,MPDynamics2,MPDynamics3} rely on the nearest-neighbor heuristic to guide the motion-tree expansion. More recent approaches leverage machine learning to improve the exploration \cite{ML8,ML11,ML7,NeuralRRTStar}.
KPIECE~\cite{KPIECE} uses interior and exterior cells, while GUST relies on discrete search over a decomposition, obtained via a triangulation, grid, or roadmap, to guide the motion-tree expansion \cite{GUST,PlakuClearance,PlakuRAL17superfacets}. There are also approaches that provide asymptotically optimal solutions (under certain assumptions) \cite{MPDynamics1,RRTStar,RRTStar2,RRTStar3,RRTStarDynamics,gammell2021asymptotically}, but they are often much slower than their suboptimal counterparts, making them unsuitable for replanning.



\paragraph{Navigation in a Partial Map}
Navigation under uncertainty is a core robot capability~\cite{Kaelbling1998pomdps,Littman1995pomdps}.
So as to mitigate computational challenges, many approaches in this domain optimistically assume unknown space is free, replanning when obstacles are revealed.
\replaced{Many path and motion-planning algorithms are problematically slow in this domain, as they must replan from scratch whenever the map changes.}{Though many path and motion-planning algorithms are applicable in this domain, they can be problematically slow or inefficient, as they must replan from scratch whenever the map changes.}
\replaced{Incremental replanning strategies reuse information between plans to accelerate grid-based replanning~\cite{koenig2002dstarlite,likhachev2005adastar}, yet most are applicable only for grid-based plans and are not well-suited for generating dynamically-feasible trajectories for nonholonomic vehicles.}{There exists a family of incremental replanning strategies that reuse information between plans to accelerate grid-based replanning, including D* Lite \cite{koenig2002dstarlite} and Anytime Dynamic A* \cite{likhachev2005adastar}; applicable only to grid-based plans, such approaches are not well-suited for generating dynamically-feasible trajectories for nonholonomic vehicles.}
Other strategies \cite{bohlin2000lazyprm,yang2022FARplanner} seek to quickly update sparse roadmaps as space is revealed, yet are similarly not designed with dynamics in mind\replaced{.}{ and so are not directly applicable for motion planning for arbitrary robots, particularly in tightly constrained environments.}
Several recent approaches \cite{adiyatov2017novel,yuan2020efficient,dai2021long} seek to reuse planning information for fast, sampling-based motion planning in dynamic environments; focused mostly on localized changes to the map, they are unlikely to be effective for long-horizon navigation in topologically rich environments, such as ours.

Instead, our work integrates multiple strategies to ensure that dynamically-feasible planning is fast, performant, and reliable despite partial observability.
\deleted{: (i) using discrete search over a fast-built graph based on adaptive subdivisions to guide sampling-based trajectory expansion, (ii) reusing previous plans to improve speed and reduce oscillatory behavior, and (iii) increasing obstacle clearance to improve robustness.}

\section{Problem Formulation}
\label{sec:problem-formulation}

This section defines the robot model and its motions, the world in which the robot operates, the sensor model, and the problem of planning collision-free and dynamically-feasible trajectories to reach the goal in an unknown environment.

\begin{figure}[t]
    \vspace{1em}
  \centering
  \includegraphics[width=0.3\columnwidth]{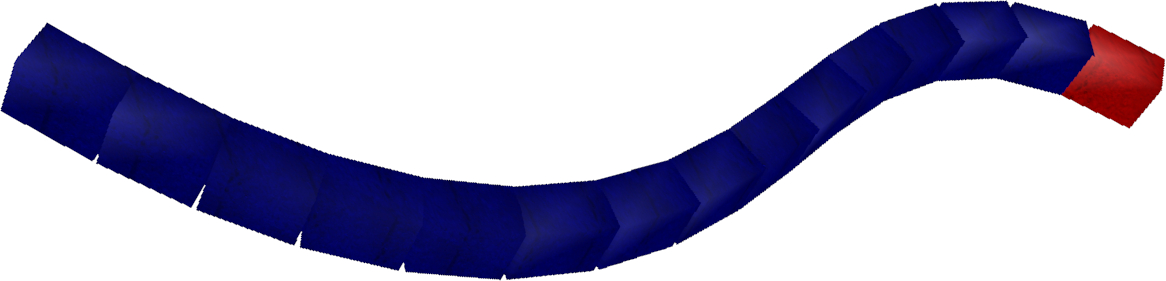}
  \caption{\textbf{The snake robot model used in the experiments.} In experiments, the number of links in the snake is varied, between 2 and 10, to control the difficulty of dynamically-feasible motion planning.}
  \label{fig:Robots}
  \vspace{-1em}
\end{figure}

\paragraph{Robot Model} 
The robot $\Robot =
 \Pair{\Shape, \StateSpace, \CtrlSpace, \MotionEqs}$ is modeled by its
 shape $\Shape$, state space $\StateSpace$, control space
 $\CtrlSpace$, and dynamics expressed as differential equations $\MotionEqs : \StateSpace \times \CtrlSpace \rightarrow \dot{\StateSpace}$. The experiments, as shown in Fig.~\ref{fig:Robots}, use a snake-like robot, modeled as a car pulling $N$ trailers~\cite{book:LaValle}. The snake head and each link are rectangular with $L=1m$ length and $W=0.6m$ width, connected with a $H=0.01m$ hitch distance. 
  The state $s = (x, y, v, \psi, \theta_0, \theta_1, \ldots, \theta_N)$ defines the position $(x, y)$, velocity $v$ ($|v|\leq 2m/s$), steering angle $\psi$ ($|\psi|\leq 1.5\Var{rad}$), head orientation $\theta_0$, and orientation $\theta_i$ for each link. The control $u = (u_a, u_\omega)$ defines the acceleration $u_a$ ($|u_a|\leq 2m/s^2$) and the steering rate $u_\omega$ ($|u_\omega|\leq 3\Var{rad}/s$). 
  The dynamics $\MotionEqs$ are defined as 
\begin{eqnarray}
\dot{x} = v \cos(\theta_0) \cos(\psi),\ \  
\dot{y} = v \sin(\theta_0) \cos(\psi), \nonumber\\[2pt]
\dot{\theta_0} = v \sin(\psi) / L, \ \ 
\dot{v} = u_a,\ \  \dot{\psi} = u_\omega,\nonumber\\[2pt]
\forall i \in \{1, \ldots, N\}:\nonumber\\[-2pt]
\dot{\theta_i} =
\frac{v}{H}(\sin(\theta_{i-1})-\sin(\theta_0)) \mbox{$\prod_{j=1}^{i-1}$}
\cos(\theta_{j-1}-\theta_j).
\end{eqnarray}

 \paragraph{Dynamically-Feasible Trajectories}
 When a control $u \in \CtrlSpace$  is applied to a state $s\in \StateSpace$, the robot moves to a new state $s_\Var{new} \in \StateSpace$ according to its dynamics. A function $\Simulate(s, u, f, dt)$ computes $s_\Var{new}$  by numerically integrating the motion equations $\MotionEqs$ for a time step $dt$, i.e.,
 \begin{equation}
     s_\Var{new} \leftarrow \Simulate(s, u, f, dt).
 \end{equation} 
To obtain a dynamically-feasible trajectory $\Traj : \{0, \ldots, \ell\} \rightarrow \StateSpace$, a sequence of controls $\Pair{u_0,
  \ldots, u_{\ell - 1}}$ is applied in succession. In this way, $\Traj(0) \leftarrow s$ and $\forall i \in \{1, \ldots, \ell\}:$
 \begin{equation}
  \Traj(i+1) \leftarrow \Simulate(\Traj(i), u_i, f, dt).
  \label{eqn:Traj}
\end{equation}

\paragraph{World and Sensor Model}
\deleted{The world $\World$ contains obstacles  within its boundaries, which are not known to the robot. The robot is equipped with a planar laser scanner, which allows it to observe nearby obstacles limited by range  and occlusion. Specifically, $\Sensor(\World, x, y, r)$ computes the free and occupied areas of $\World$ that are visible from $(x, y)$ and no more than $r$ units away, where $r \in \R^{>0}$ is the sensor range. }
The world $\World$ has unknown obstacles within its boundaries. The robot uses a planar laser scanner to observe nearby obstacles, limited by its range $r \in \R^{>0}$ and occlusion. Specifically, as shown in Fig.~\ref{fig:MainScene}, $\Sensor(\World, x, y, r)$ computes visible free and occupied areas of $\World$ from $(x, y)$ within a distance of $r$.
The robot uses the sensor readings to build a model of the world, as a high-resolution occupancy grid $\SensorGrid$,  on which it can plan its motions. Initially, all cells in $\SensorGrid$ are unknown. When the robot moves to $(x, y)$, $\SensorGrid$ is updated using the free and occupied cells computed by $\Sensor(\World, x, y, r)$.  

$\Function{collision}(\SensorGrid, s)$ checks whether a state $s \in \StateSpace$ is invalid. A state $s$ is invalid if a value is outside the designated bounds or if there is collision with known obstacles or between non-consecutive links, when the robot is placed at position $(x, y)$ and orientations $\theta_0, \ldots, \theta_N$ defined by $s$.

\paragraph{Overall Problem} Our framework receives as input the robot model $\Robot =
 \Pair{\Shape, \StateSpace, \CtrlSpace, \MotionEqs}$, a sensor model $\Sensor$, an initial state $s_\Var{init} \in \StateSpace$, and a goal $\Goal$. The initial state and the goal are within the world boundaries, which are known to the framework. The obstacles contained in $\World$, however, are not known to the framework. The objective is to compute controls so that the resulting dynamically-feasible trajectories enable  the robot to reach the goal while avoiding collisions. Our framework seeks to reduce the overall planning time and the distance traveled by the robot.
\deleted{Fig.~\ref{fig:Decomp} shows snapshots of our framework in action. Videos of solutions can be found in the supplementary material or at https://tinyurl.com/47ct55s6.}
\section{Method}
\label{sec:method}

\begin{figure*}
\vspace{1em}
\centering
\includegraphics[width=0.24\textwidth]{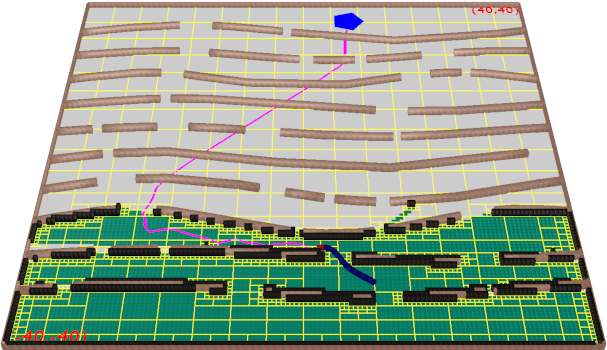}
\includegraphics[width=0.24\textwidth]{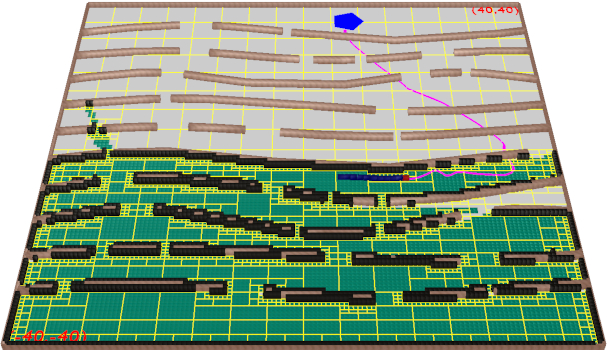}
\includegraphics[width=0.24\textwidth]{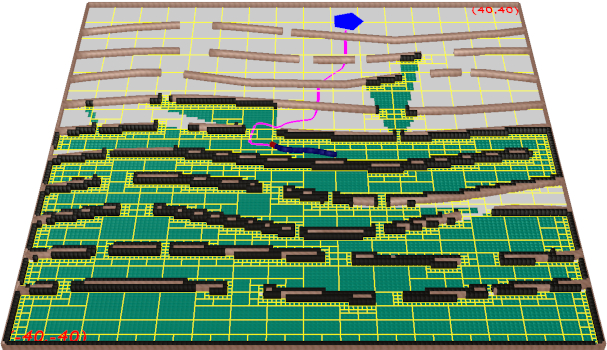}
\includegraphics[width=0.24\textwidth]{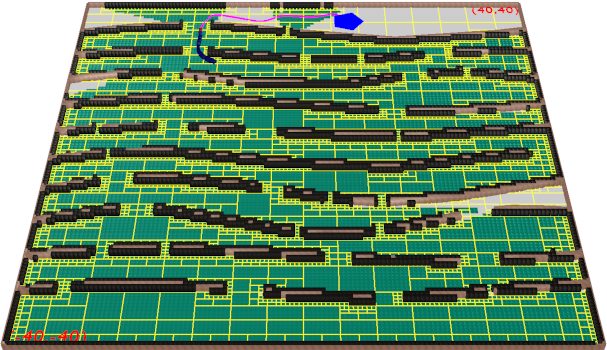}
\caption{\textbf{Snapshots of our framework in action.} The free areas in $\SensorGrid$ are shown in green, the occupied areas in black, and the remaining areas represent unknown space. The robot is shown at its current state (robot in dark blue with its head in red). The planned trajectory from the current state to the goal is shown in magenta. The subdivision is shown in yellow. Figures are better viewed in color and on screen.}
\label{fig:Decomp}
\vspace{-1em}
\end{figure*}

Our framework has an execution module (EM) and a planning module (PM).  The execution module carries out the planned trajectory incrementally, while the planning module is responsible for generating collision-free and dynamically-feasible trajectories through the partial map. The planning module is invoked by the execution module as necessary when new obstacles are detected. Fig.~\ref{fig:Decomp} shows snapshots of our framework in action. Further details are provided below.

\begin{algorithm}[t]
\caption{Execution Module (EM)}
\label{alg:ExecutionModule}
\begin{algorithmic}[1]
\INPUT{$\Robot$: robot model; $\Sensor$: sensor model; $\World$: world model, but only the boundaries are known to the framework, obstacles are discovered through sensing; $s_\Var{init}$: initial state; $\Goal$: goal region; $\Var{tmax}_\Var{em}$: maximum runtime for the execution module; 
$\Var{maxNrFailsAllowed}$: maximum number of consecutive failures for the planning module before giving up}; $r$: sensor range
\OUTPUT{executed trajectory $\Traj_\Var{exe}$, which is built by invoking the planning module along the way to the goal}\\[1mm]
\hrule
\setcounter{ALC@line}{0}
\vspace*{0.5mm}
\STATE $\Function{initialize}(\SensorGrid)$ \CodeComment{occupancy grid, initialized to unknown}
\STATE $\Traj_\Var{exe}.\Function{insert}(s_\Var{init})$ \CodeComment{execution trajectory starts from initial state}
\STATE $\Traj_\Var{mp} \leftarrow \emptyset$ \CodeComment{planned trajectory is initially empty}
\STATE $\Var{pos} \leftarrow 0$ \CodeComment{index to track remaining states in $\Traj_\Var{mp}$}
\STATE $\Var{replan} \leftarrow \True; d \leftarrow 0$ \CodeComment{when to invoke the planning module}
\WHILE{$\neg\Function{reached}(\Traj_\Var{exe}(\Var{end}), \Goal) \wedge \Function{time}() < \Var{tmax}_\Var{em}$}
\STATE $s_\Var{curr} \leftarrow \Traj_\Var{exe}(\Var{end})$ \CodeComment{get last state of execution trajectory}
\STATE $\Function{UpdateRobotState}(\Robot, s_\Var{curr})$ \CodeComment{move robot to state}
\STATE $\Pair{\Var{cells}_\Var{free}, \Var{cells}_\Var{occ}} \leftarrow \Sensor(\World, s_\Var{curr}.x, s_\Var{curr}.y, r)$
\STATE \textbf{if} {$\exists c\in\Var{cells}_\Var{occ}$: $\SensorGrid(c) = \Constant{unknown}$   \textbf{or} \\ $\Var{pos} \geq |\Traj_\Var{mp}|$}
       \textbf{then} $\Var{replan} \leftarrow \True$
\STATE $\Function{update}(\SensorGrid, \Var{cells}_\Var{free}, \Var{cells}_\Var{occ})$
\STATE $\Var{nrFails} \leftarrow 0$ \CodeComment{keep track of consecutive planner failures}
\WHILE{$\Var{replan}$}
\STATE $\Traj_\Var{prev} \leftarrow \Pair{\Traj_\Var{mp}(\Var{pos}), \ldots, \Traj_\Var{mp}(\Var{end})}$
\STATE $\Traj_\Var{mp}\!\! \leftarrow\! \Function{PlanningModule}(\World_\Var{sensed},\! \Robot,\! s_\Var{curr},\! \Goal,\! \Traj_\Var{prev}\!)$
\STATE \textbf{if} $|\Traj_\Var{mp}| \leq 1$ \textbf{then} ++$\Var{nrFails}$ \CodeComment{planner failed}
\STATE \textbf{else} $\{ \Var{replan} \leftarrow \False; \Var{pos} \leftarrow 0;\}$ 
\IF{$\Var{nrFails} \geq \Var{maxNrFailsAllowed}$}
\STATE \textbf{return} $\Traj_\Var{exe}$ \CodeComment{give up after several consecutive failures}
\ENDIF
\ENDWHILE
\STATE $\Traj_\Var{exe}.\Function{insert}(\Traj_\Var{mp}(\Var{pos}))$; ++$\Var{pos}$
\ENDWHILE
\STATE \textbf{return} $\Traj_\Var{exe}$
\end{algorithmic}
 \end{algorithm}

\subsection{Execution Module}
\label{sec:ExecutionModule}
Pseudocode for the execution module (EM) is shown in Alg.~\ref{alg:ExecutionModule}. The module starts by marking all the space in the high-resolution occupancy grid $\SensorGrid$ as unknown 
(Alg.~\ref{alg:ExecutionModule}:1). It then continues until either the robot successfully reaches its goal or reaches a predetermined runtime limit.

The execution module keeps track of two trajectories: $\Traj_\Var{exe}$ and $\Traj_\Var{mp}$. The trajectory $\Traj_\Var{exe}$ 
records the robot's motion from its initial state to its current location. The planning module computes the trajectory $\Traj_\Var{mp}$, which indicates the sequence of states that the robot should follow to reach the goal. Initially, $\Traj_\Var{exe}$ contains only $s_\Var{init}$, and $\Traj_\Var{mp}$ is empty since the planning module has not been invoked yet (Alg.~\ref{alg:ExecutionModule}:2--3).

During each iteration, the robot uses its sensor to detect the free and occupied space near its current state, constrained by both the sensor range and potential occlusions 
(Alg.~\ref{alg:ExecutionModule}:7--9). The detected space is used to update $\SensorGrid$  (Alg.~\ref{alg:ExecutionModule}:11). 
 
 If the sensor detects an area that was previously marked as unknown but is now identified as occupied, the execution module invokes the planning module to plan a new collision-free and dynamically-feasible trajectory from the current state to the goal. This is because the newly detected obstacle may collide with the previously planned trajectory.  The planning module is also invoked when the robot reaches the end of the trajectory $\Traj_\Var{mp}$ (Alg.~\ref{alg:ExecutionModule}:10). 
 
 To enhance the planning efficiency, the planning module is provided with the remaining states in $\Traj_\Var{mp}$ as input (Alg.~\ref{alg:ExecutionModule}:14--15), which it can use to guide the motion-tree expansion to more efficiently generate a new trajectory towards the goal. If the planner is unable to generate a collision-free and dynamically-feasible trajectory that leads to the goal, it returns the motion-tree trajectory that is closest to the goal. When the planner completely fails, meaning it cannot generate a trajectory that extends beyond the current state, it is invoked again. However, after multiple consecutive failures, the execution module abandons the planning process (Alg.~\ref{alg:ExecutionModule}:18--19), and returns unsuccessfully.

When the planner succeeds, the execution module progresses to the subsequent state in the planned trajectory (Alg.~\ref{alg:ExecutionModule}:20). This process of following the planned trajectory and re-planning when new obstacles are detected is repeated until the robot reaches its goal or exceeds the runtime limit.

\subsection{Planning Module}
\label{sec:PlanningModule}

Pseudocode for the planning module (PM) is shown in Alg.~\ref{alg:PlanningModule}. 
Given the (partial) occupancy grid $\SensorGrid$, which represents the sensed world as free, occupied, and unknown cells, the planning module seeks to compute a dynamically-feasible trajectory from the current state $s_\Var{curr}$ to the goal $\Goal$ that avoids occupied cells. The planning module optimistically assumes unknown space is obstacle-free and so allows the robot to move through it. If new obstacles are detected in unknown space, the execution module prompts the planning module to generate a new obstacle-avoiding trajectory.
\deleted{The planning module allows the robot to move through unknown space, as it optimistically assumes that this space is free of obstacles. However, if new obstacles are detected in the previously unknown space, the execution module will invoke the planning module again to generate a new trajectory that avoids the newly detected obstacles.}

The planning module comprises two layers: (i) the high-level discrete layer determines general directions towards the goal; and (ii) the low-level continuous layer expands a motion tree along the identified directions.
\deleted{The planning module is composed of two layers: (i) the high-level discrete layer, which is responsible for determining general directions towards the goal; and (ii) the low-level continuous layer, which expands a motion tree along the general directions identified by the high-level layer. }

\subsubsection{High-Level Discrete Layer Based on Grid Subdivision} The high-level layer operates on a subdivision of the free and unknown space in $\SensorGrid$ (Alg.~\ref{alg:PlanningModule}:1). This involves placing a coarse-resolution grid over the boundaries, and subdividing each cell until its area is either no larger than the cells in $\SensorGrid$, or it does not contain any occupied cells from $\SensorGrid$. Fig.~\ref{fig:Decomp} shows some  examples.

The subdivision is preferred over operating directly on  $\SensorGrid$ because it reduces the computational cost of searching for paths to the goal. In fact, the high resolution of $\SensorGrid$, which is required for an accurate representation of the environment, imposes significant computational costs on the search. In contrast, the subdivision approach offers adaptive resolution, allowing the high-level layer to work with a coarser grid in areas where there are no obstacles, and a finer grid in areas where obstacles are present. This reduces the overall computational cost, while still providing an accurate representation of the environment.

We utilize Dijkstra's single-source shortest-path algorithm to calculate the shortest paths from the goal to each region in the subdivision (Alg.~\ref{alg:PlanningModule}:3). To speed up the process, we employ radix heaps. Moreover, each region $r$ retains its path to the goal,  denoted as $r.\Var{pathToGoal}$, which guides the low-level continuous layer during the motion-tree expansions.

 \begin{algorithm}[t]
\caption{Planning Module (PM)}
\label{alg:PlanningModule}
\begin{algorithmic}[1]
\INPUT{$\Robot$: robot model; $\SensorGrid$: occupancy grid representing the sensed world (free, occupied, unknown); $s_\Var{curr}$: start state; $\Goal$: goal region; $\Traj_\Var{prev}$: sequence of states representing previous planned trajectory; $\Var{tmax}_\Var{mp}$: maximum runtime for the planning module}
\OUTPUT{planned collision-free and dynamically-feasible trajectory that reaches $\Goal$; if not, return best trajectory}\\[1mm]
\hrule
\setcounter{ALC@line}{0}
\vspace*{0.5mm}
\STATE $\Subdiv \leftarrow \Function{subdivision}(\SensorGrid)$
\STATE $\Function{ClearancesFromOccupied}(\Subdiv)$
\STATE $\Function{PathsToGoal}(\Subdiv, \Goal)$
\STATE $\Tree \leftarrow \Function{InitializeMotionTree}(s_\Var{curr}, \Traj_\Var{prev})$
\STATE $ \mtv_\Var{best} \leftarrow \Tree.\Var{root}$
\WHILE{$\Function{time}() < \Var{tmax}_\Var{mp}$}
\STATE $r \leftarrow \Function{SelectRegion}(\Subdiv)$
\STATE $\Var{groups} \leftarrow \Pair{\Var{group}_i : 0 \leq i < |r.\Var{pathToGoal}|}$
\STATE $\Var{group}_0.\Var{nodes} \leftarrow \{\Function{SelectNode}(r.\Var{nodes})\}; i \leftarrow 0$
\FOR{$i  < |r.\Var{pathToGoal}|$ \textbf{and} several iterations}
\STATE $\Var{group}_j \leftarrow \Function{SelectGroup}(\Var{group}_0, \ldots, \Var{group}_i)$
\STATE $p_\Var{target} \leftarrow \Function{SampleTargetPosition}(\Var{group}_j)$
\STATE $\mtv \leftarrow \Function{SelectNode}(\Var{group}_j)$
\FOR{several iterations}
\STATE $u \leftarrow \Function{controller}(\mtv.s, p_\Var{target})$
\STATE $s_\Var{new} \leftarrow \Simulate(\mtv.s, u, \MotionEqs, dt)$
\STATE \textbf{if} $\Function{collision}(\SensorGrid, s_\Var{new})$ \textbf{then break}
\STATE $\mtv_\Var{new} \leftarrow \Function{NewNode}(); \Tree.\Function{insert}(\mtv_\Var{new})$
\STATE $\mtv_\Var{new}.\Pair{s, u, \Var{parent}} \leftarrow \Pair{s_\Var{new}, u, \mtv}$
\STATE \textbf{if} $\Function{reached}(s_\Var{new}, \Goal)$ \textbf{then return} $\Traj_\Tree(\mtv_\Var{new})$
\STATE $\Var{rnew} \leftarrow \Function{LocateRegion}(\Subdiv, s_\Var{new})$
\STATE $\Var{rnew}.\Var{nodes}.\Function{insert}(\mtv_\Var{new}); \mtv_\Var{new}.r \leftarrow \Var{rnew}$
\STATE \textbf{if} $\Function{far}(s_\Var{new}, r.\Var{pathToGoal}(j))$ \textbf{then break} 
\STATE \textbf{if} $\Function{reached}(s_\Var{new}, r.\Var{pathToGoal}(j))$ \textbf{then} ++$j$
\STATE $\Var{group}_j.\Var{nodes}.\Function{insert}(\mtv_\Var{new})$
\STATE $i \leftarrow \max(i, j)$
\STATE \textbf{if} $\mtv_\Var{best}.r.\Var{pathToGoal}.\Var{cost} >$\\\hspace*{2.9mm} $\Var{rnew}.\Var{pathToGoal}.\Var{cost}$ \textbf{then} $\mtv_\Var{best} \leftarrow \mtv_\Var{new}$
\STATE $\mtv \leftarrow \mtv_\Var{new}$
\ENDFOR
\ENDFOR
\ENDWHILE
\STATE \textbf{return} $\Traj_\Tree(\mtv_\Var{best})$

\end{algorithmic}
 \end{algorithm}
 
When computing the shortest paths, the cost of traveling between two adjacent regions, $r_i$ and $r_j$, is not exclusively based on the Euclidean distance between their centers. Instead, it also accounts for their respective clearances, $\Var{clear}(r_i)$ and $\Var{clear}(r_j)$, from obstacles. Specifically, 
\begin{equation}
    \Function{cost}(r_i, r_j) = \frac{||\Var{center}(r_i) - \Var{center}(r_j)||_2}{ (\min(\Var{clear}(r_i), \Var{clear}(r_j), \Var{cmax}))^\alpha}.
\end{equation}
High values of $\alpha$ ($\alpha=6$ in the experiments) emphasize clearance over distance. 
Paths with ample clearance  are less prone to collisions when new obstacles are detected. However, the discovery of a new obstacle may cause the clearance to decrease, leading to significant changes in the paths to the goal from certain regions. To prevent such abrupt modifications, we introduce $\Var{cmax}$ (set to $4m$ in the experiments), which sets the maximum allowable clearance.

The clearances (Alg.~\ref{alg:PlanningModule}:2) are efficiently computed based on brush-fire search \cite{book:MP}. First, all the occupied regions are inserted into a priority queue with clearance of $0$. Next, any non-occupied region on the boundary is added to the queue with a clearance equal to its distance from the boundary. While the queue is not empty, we extract the region $r$ with the minimum clearance. Each adjacent region $r_\Var{adj}$ whose clearance has not yet been defined is inserted into the queue with a clearance of  $\Var{clear}(r) + ||\Var{center}(r) - \Var{center}(r_\Var{adj})||_2$.

\begin{figure*}
\vspace{1em}
\centering
\begin{tabular}{ccc}
\includegraphics[width=0.32\textwidth]{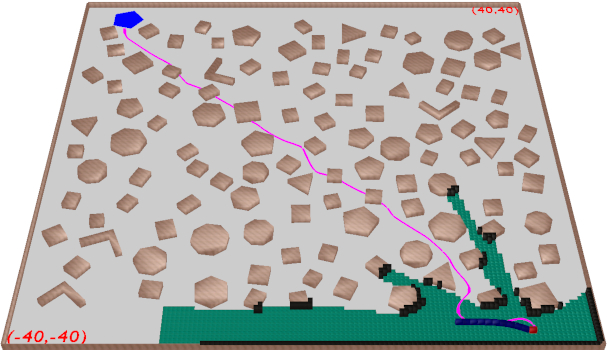}\hspace*{-4mm} & 
\includegraphics[width=0.32\textwidth]{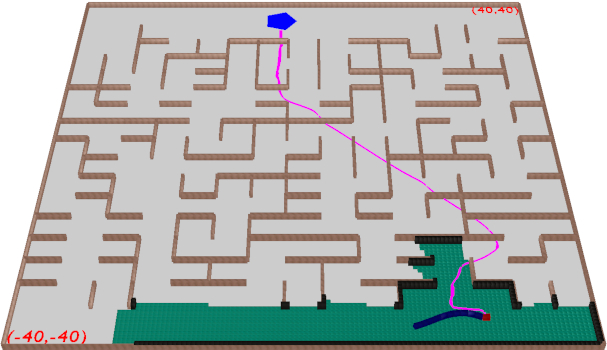}\hspace*{-4mm} &
\includegraphics[width=0.32\textwidth]{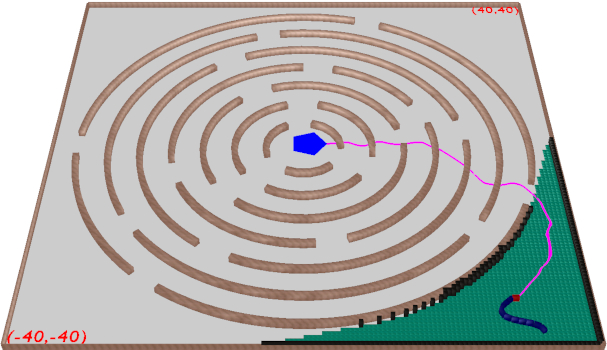}\\[-1mm]
{\footnotesize{environment type 2}} &
{\footnotesize{environment type 3}} &
{\footnotesize{environment type 4}}
\end{tabular}\\
\begin{tabular}{cc}
\includegraphics[width=0.35\textwidth]{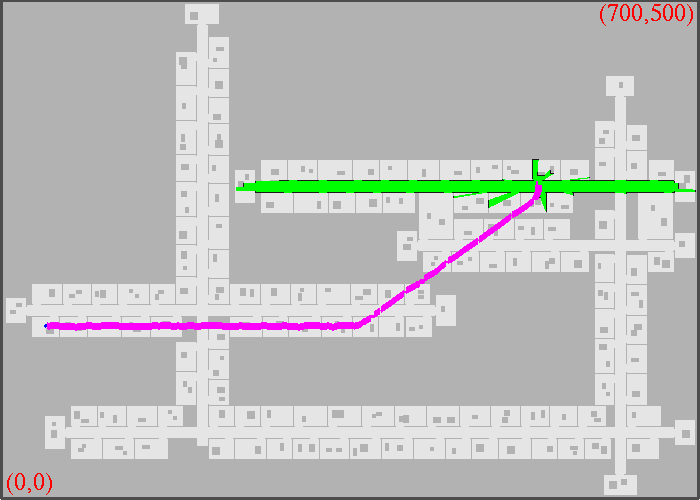}\hspace*{-4mm} &
\includegraphics[width=0.199\textwidth]{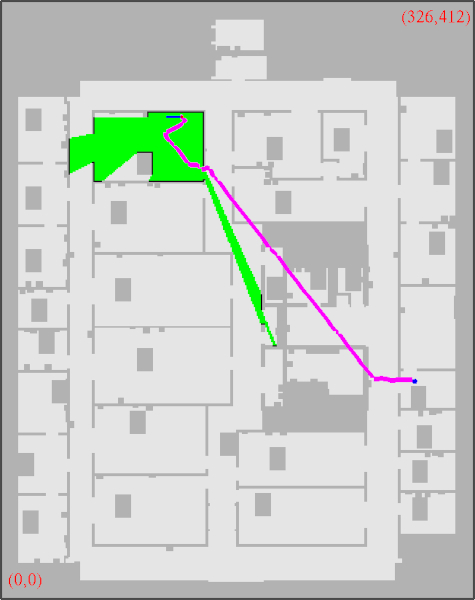}\\[-1mm]
{\footnotesize{environment type 5}} &
{\footnotesize{environment type 6}}
\end{tabular}
\caption{\textbf{Example maps of environment types 2--6 used in the experiments.} (Environment type 1 is shown in Fig.~\ref{fig:MainScene}). 
Each environment type has multiple levels of difficulty, with at least 30 instances for each level. Figures show instances with the highest difficulty level for environment types 1--5 and the second-highest for environment type 6. Sensor readings are shown in green (free) and black (occupied) within a sensor radius of 50m from the robot's initial state. 
The scene coordinates are in meters. The planned trajectory from the initial state to the goal is shown in magenta.}
\label{fig:OtherScenes}
\vspace{-1em}
\end{figure*}

\subsubsection{Low-Level Continuous Layer Based on Guided Motion-Tree Expansion}

The continuous layer of the planning module expands a motion tree, denoted by $\Tree$, by adding collision-free and dynamically-feasible trajectories as branches.
These trajectories are generated by applying a controller that guides the robot from a state in $\Tree$ to a  target position. 

\paragraph{Motion-Tree Representation} Each node $\mtv \in \Tree$ has four fields: $\Pair{s, u, \Var{parent}, r}$, which correspond to its state, control, parent, and subdivision region, respectively. The node's state, $\mtv.s$, is obtained by applying the control $\mtv.u$ to the parent state:  i.e., $\mtv.s \leftarrow \Simulate(\mtv.\Var{parent}.s, \mtv.u, \MotionEqs, dt)$. The construction ensures that $\mtv.s$ does not collide with any occupied cell in $\SensorGrid$. The node $\mtv$ also records the subdivision region, $\mtv.r$, that contains its position $(\mtv.s.x, \mtv.s.y)$.

\paragraph{Retrieving the Best Trajectory}
A solution is found found when a new node $\mtv_\Var{new}$ reaches the goal $\Goal$. The solution then corresponds to the sequence of states from the root of  $\Tree$ to $\mtv_\Var{new}$, denoted as $\Traj_\Tree(\mtv_\Var{new})$. If $\Goal$ is not reached, the framework keeps track of the best trajectory to $\Goal$. The node $\mtv_\Var{best}$ is initially set as the root of $\Tree$. When a new node, $\mtv_\Var{new}$, is added to $\Tree$, the cost of the path to $\Goal$ from its region, given by $\mtv_\Var{new}.r.\Var{pathToGoal}.\Var{cost}$, is compared with the cost of $\mtv_\Var{best}$'s path to $\Goal$. If the cost of the new node is lower, then $\mtv_\Var{best}$ is updated to the new node.

\paragraph{Motion-Tree Initialization by Leveraging the Prior Solution} Motion-tree initialization starts by setting the current state $s_\Var{curr}$ as the root of $\Tree$. If the prior solution $\Traj_\Var{prev}$ is available, the states in $\Traj_\Var{mp}$ are processed one by one (Alg.~\ref{alg:PlanningModule}:4). For each state, if it is not in collision with the occupied cells of $\World_\Var{sensed}$, it is added as a new node to $\Tree$, with the previous state in $\Traj_\Var{prev}$ serving as its parent. Otherwise, the initialization stops.

\paragraph{Guided Motion-Tree Expansion} The motion-tree expansion leverages the path-to-goal for the subdivision regions. Specifically, let $\Gamma$ denote the regions $\Delta$ that have been reached by $\Tree$. At each iteration, the region with the maximum weight is selected for expansion from $\Gamma$ (Alg.~\ref{alg:PlanningModule}:7). The weight of region $r$ is defined as
\begin{equation}
    r.w = \beta^{r.\Var{nsel}}/ r.\Var{pathToGoal}.\Var{cost},    
\end{equation}
where $r.\Var{nsel}$ denotes the number of previous selections and $0 < \beta < 1$. This weighting scheme prioritizes regions that have low-cost paths to the goal. However, we do not want to repeatedly select the same region when expansion attempts fail due to obstacles or robot dynamics. To prevent this, we apply a penalty factor, $\beta$,  after each selection. This ensures expansions from new regions in the subdivision.

After selecting a region $r$, the objective is to expand $\Tree$ along $r.\Var{pathToGoal}$. \deleted{This requires the motion-tree branches to remain close to $r.\Var{pathToGoal}$.}
To simplify the process, we create a group, denoted as $\Var{group}_i$, for each index in  $r.\Var{pathToGoal}$ (Alg.~\ref{alg:PlanningModule}:8). Each $\Var{group}_i$ contains nodes from $\Tree$ seeking to reach the $i$-th region in $r.\Var{pathToGoal}$. Once a node reaches region $r.\Var{pathToGoal}(i)$, $\Var{group}_{i+1}$ becomes available. This is repeated for several iterations or until the end of $r.\Var{pathToGoal}$ is reached.

To start, we randomly choose a node from the ones that have reached $r$, and add it to $\Var{group}_0$ (Alg.~\ref{alg:ExecutionModule}:9). We also keep track of the progress made in following $r.\Var{pathToGoal}$ using an index $i$. During each iteration, we select the group with the highest weight from the set $\Var{group}_0, \ldots, \Var{group}_i$. The weight of $\Var{group}_j$ is defined as
\begin{equation}
    \Var{group}_j.w = 2^j \beta^{\Var{group}_j.\Var{nsel}}.    
\end{equation}
The term $2^j$ gives priority to groups closer to the end of $r.\Var{pathToGoal}$, while a penalty factor ensures that the same group is not selected indefinitely.

Let $\Var{group}_j$ denote the selected group (Alg.~\ref{alg:PlanningModule}:11). Since the objective is to reach $r.\Var{pathToGoal}(j)$, a target position $p_\Var{target}$ is sampled inside $r.\Var{pathToGoal}(j)$ (Alg.~\ref{alg:PlanningModule}:12). Next, the closest node $\mtv$ to $p_\Var{target}$ from $\Var{group}_j.\Var{nodes}$ is selected with the objective of expanding it toward $p_\Var{target}$ (Alg.~\ref{alg:PlanningModule}:13).

To expand $\mtv$ towards $p_\Var{target}$, a PID controller is used to compute the sequence of controls that steer the robot towards $p_\Var{target}$, for example, by turning the wheels (Alg.\ref{alg:ExecutionModule}:15). If the new state $s_\Var{new}$ is in collision, the expansion towards the target stops (Alg.\ref{alg:ExecutionModule}:17). Otherwise, a new node is added to $\Tree$ (Alg.~\ref{alg:ExecutionModule}:18-19). If the new node reaches the goal, the planner terminates successfully. If the new node is far from $r.\Var{pathToGoal}(j)$, then the expansion towards the target stops. However, if the new node reaches $r.\Var{pathToGoal}(j)$, the next group $\Var{group}_{j+1}$ is made available for selection, and the new node is added to this group.

In this manner, the planning module incrementally expands the motion tree along the paths to the goal that are associated with the regions in the subdivision. When progress along a path slows down, the planner searches for alternative paths from different regions. By doing so, the planning module is able to efficiently generate collision-free and dynamically-feasible trajectories to the goal, as demonstrated by the experimental results.

\begin{figure*}
    \vspace{1em}
    \centering
    \includegraphics[width=\textwidth]{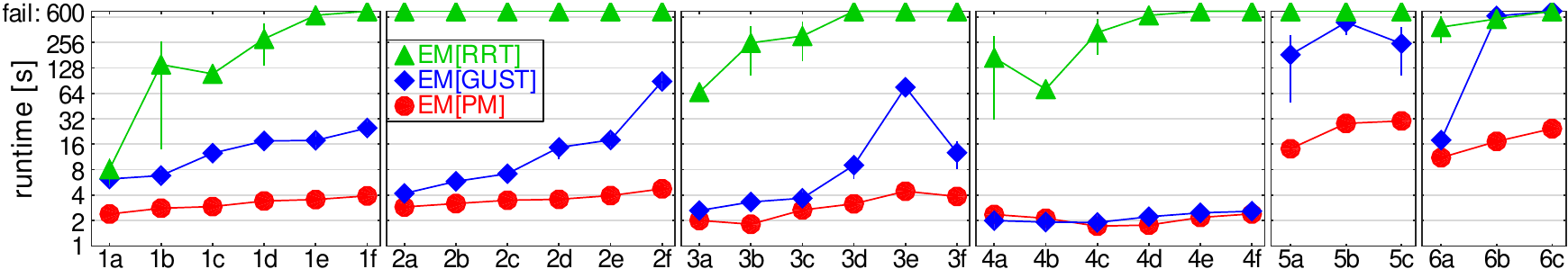}\\
    \includegraphics[width=\textwidth]{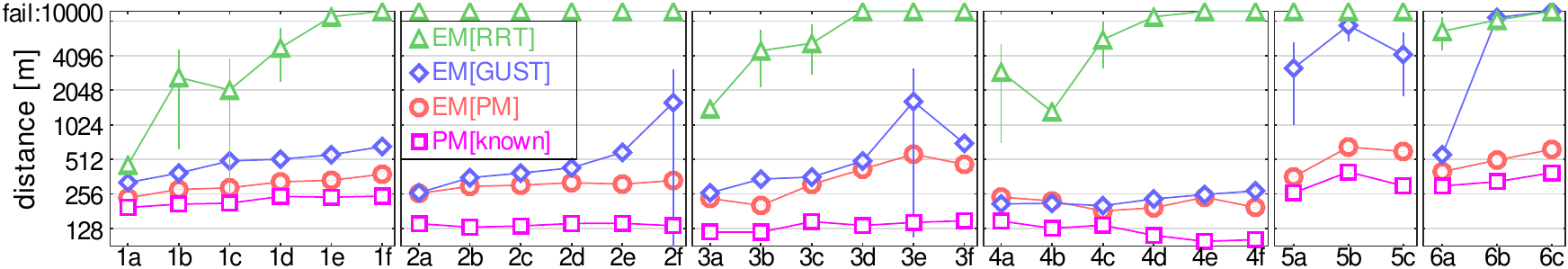}
    \caption{\textbf{Results when using the execution module (EM) with different motion planners.} The graph has a logarithmic $y$-axis and an $x$-axis showing environment type and difficulty level, e.g., $3b$ means the third environment type and second difficulty level. The results are for a $5$-link snake model with $50m$ sensor range. PM[known] shows the distance results of the motion planner when the environment is fully known (mean runtime $<1s$ for each scene).}

    \label{fig:ResTimeAndDist}
\end{figure*}

\section{Experiments and Results}
Experiments are conducted in simulation using a snake robot model with nonlinear dynamics.
We use over a thousand virtual environments, both structured and unstructured. To successfully navigate these environments, the robot must rely on information gathered from its sensor to traverse unknown spaces, avoid various obstacles, and maneuver through narrow passages to ultimately reach its goal.

\subsection{Experimental Setup}

\subsubsection{Methods used for Comparisons} We use our execution module (Alg.\ref{alg:ExecutionModule}) together with the planning module (Alg.\ref{alg:PlanningModule}) to conduct experiments, denoted as EM[PM]. Additionally, the performance of the execution is evaluated when combined with other motion planners, namely RRT~\cite{RRT,RRTRecent1} and GUST~\cite{PlakuTRO15}, denoted as EM[RRT] and EM[GUST], respectively. RRT is selected due to its popularity as a sampling-based motion planner, while GUST is chosen for its computational efficiency in motion planning with dynamics. To ensure fair comparisons, our implementations of RRT and GUST are fine-tuned for the experiments in this paper.

\subsubsection{Environments} \deleted{Experiments are conducted in six different types of environments, as shown in Figs.~\ref{fig:MainScene} and \ref{fig:OtherScenes}. Each environment type has varying levels of difficulty, and for each type and level of difficulty, at least $30$ instances are generated.} 
We conduct experiments in six environment types (Figs.~\ref{fig:MainScene} and \ref{fig:OtherScenes}) with varying difficulty levels. For each type and difficulty level, we generate at least 30 instances. \deleted{We refer to the set of all instances corresponding to a particular environment type and level of difficulty as $\Symbol{I}_\Pair{\Var{type}, \Var{level}}$.}
Environment types 1 to 4 have dimensions of $80m\times 80m$, while environment types 5 and 6 are much larger, ranging from $15000m^2$ to $350000m^2$. This allows us to test the framework's ability to navigate different sized environments. $\SensorGrid$, the occupancy map, has $128\times128$ cells for environment types 1--4 and $\Var{dimx}\times \Var{dimy}$ for environment types 5--6, where $\Var{dimx}$ and $\Var{dimy}$ are the scene dimensions (ranging from $100$ to $850$). The coarse-grid used to initialize the adaptive subdivision has $48\times 48$ cells for environment types 1--4 and $64\times 64$ cells for types 5--6.


\paragraph{Environment Type 1}
The first environment type (Fig.~\ref{fig:MainScene}) has  wave-like obstacles separated by some distance, with randomly placed gaps of varying widths. Some gaps are narrow, making it difficult or impossible for the robot to navigate through. We use six difficulty levels, with scenes ranging from five to ten waves. For each instance, the start and goal are placed at random unoccupied locations near the bottom and top, respectively.

\paragraph{Environment Type 2}
The second type (Fig.~\ref{fig:OtherScenes}) includes randomly distributed obstacles. We adjust the level of difficulty by modifying the density of obstacles, using six levels ranging from $15\%$ to $20\%$ obstacle coverage. The start and goal for each instance are again placed at random unoccupied locations near the bottom and top, respectively.

\paragraph{Environment Type 3}
For the third type (Fig.~\ref{fig:OtherScenes}), we generate mazes using Kruskal's algorithm, with the size varied for six difficulty levels, ranging from $10\times 10$ to $15\times 15$. 
The start and goal for each instance are generated using the same procedure as environment types 1 and 2.

\paragraph{Environment Type 4}
Environment type 4 (Fig.~\ref{fig:OtherScenes}) features concentric rings with randomly placed gaps of varying widths. Six difficulty levels are created by varying the separation between the rings, ranging from $7m$ to $2m$. For each instance, the start and goal are randomly placed outside the largest ring and inside the smallest ring, respectively.

\paragraph{Environment Type 5}
Environment type 5 (Fig.~\ref{fig:OtherScenes}) emulates an office building by generating occupancy grid maps with intersecting hallways and offices/meeting rooms bordering them, with furniture-like obstructions making it difficult to see the full room without entering. Three difficulty levels are created by varying the floor dimensions to $500m \times 300m$, $600m \times 400m$, and $700m \times 500m$. The start and goal for each instance are placed at random unoccupied locations.
\deleted{The fifth environment type (Fig.~\ref{fig:OtherScenes}) is designed to emulate an office building, with occupancy grid maps generated by first placing a set number of intersecting hallways and then adding offices and meeting rooms bordering those hallways; rooms contain furniture-like obstructions, so that it is difficult in general to see the full room without entering it. We created three difficulty levels by changing the floor's dimensions to $500m \times 300m$, $600m \times 400m$, and $700m \times 500m$. The start and goal for each instance are placed at random unoccupied locations.}

\paragraph{Environment Type 6}
\deleted{The sixth type (Fig.~\ref{fig:OtherScenes}) consists of occupancy grid maps computed from a dataset of real-world floor plans of buildings around the MIT campus~\cite{whiting2007topology}.
The difficulty levels are based on the floor plan's size, with the first level containing floor plans between $93000m^2$ and $101906m^2$ (33 instances), the second level having floor plans between $120330m^2$ and $188914m^2$ (86 instances), and the third level comprising floor plans between $207414m^2$ and $282800m^2$ (59 instances). 
For each instance, the start and goal are placed at random unoccupied locations.}
The sixth type (Fig.\ref{fig:OtherScenes}) uses occupancy grid maps generated from real-world floor plans of buildings around the MIT campus\cite{whiting2007topology}. Three difficulty levels are created based on floor plan size, with 33 instances for level one (floor plans between $93000m^2$ and $101906m^2$), 86 instances for level two (floor plans between $120330m^2$ and $188914m^2$), and 59 instances for level three (floor plans between $207414m^2$ and $282800m^2$). For each instance, start and goal locations are randomly placed in unoccupied areas.

\subsubsection{Measuring Performance} The execution module is run with our planning module, RRT, and GUST on each of the $1078$ instances. 
We evaluate performance based on the environment type and difficulty level, reporting the total runtime and distance traveled by the robot. To prevent outlier effects, we compute the mean runtime and distance traveled after removing the top and bottom $20\%$ of results.
\deleted{We report the performance of EM[PM], EM[RRT], and EM[GUST] based on the environment type and level of difficulty. We report the overall runtime of the planning module and the overall distance traveled by the robot (length of $\Traj_\Var{exe}$). To avoid the impact of outliers, we calculate the mean runtime and distance traveled after discarding the worst and best $20\%$ of results.}

\subsubsection{Computing Resources} The experiments ran on HOPPER, a computing cluster provided by GMU's Office of Research Computing. Each node has 48 cores with Dell PowerEdge R640 Intel(R) Xeon(R) Gold 6240R CPU 2.40GHz. The experiments were not parallelized or multi-threaded, and each instance was executed on a single core. The code was developed in C++ and compiled with g++-9.3.0.

\begin{figure*}
    \centering
    \includegraphics[width=0.49\textwidth]{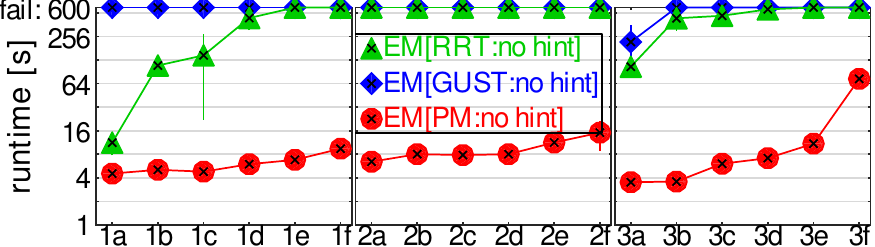}
    \includegraphics[width=0.49\textwidth]{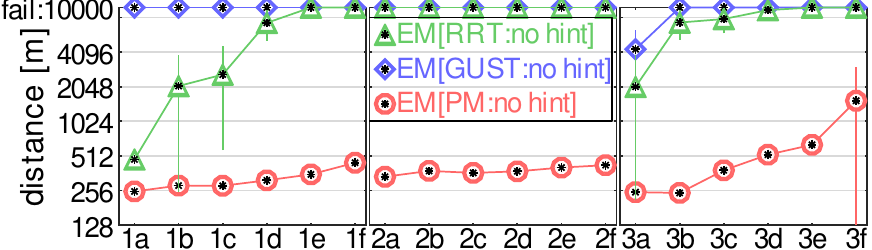}
    \caption{\textbf{Results when disabling the hint}: i.e., the execution module does not provide the prior solution as input to the motion planners.}
    \label{fig:ResNoHints}
\end{figure*}

\begin{figure*}
    \centering
    \includegraphics[width=0.49\textwidth]{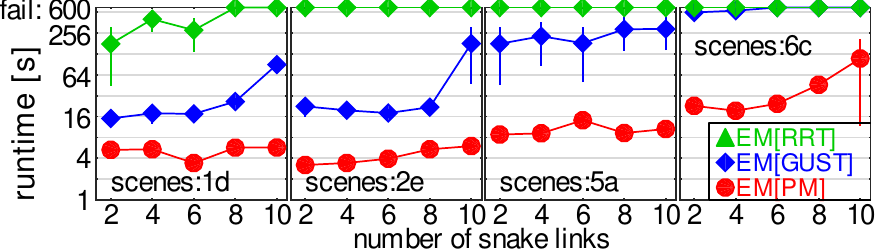}
    \includegraphics[width=0.49\textwidth]{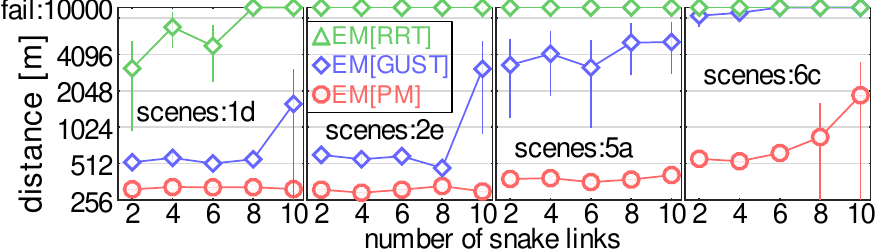}
    \caption{\textbf{Results when varying the number of snake links.}}
    \label{fig:ResLinks}
\end{figure*}

\begin{figure*}
    \centering
    \includegraphics[width=0.49\textwidth]{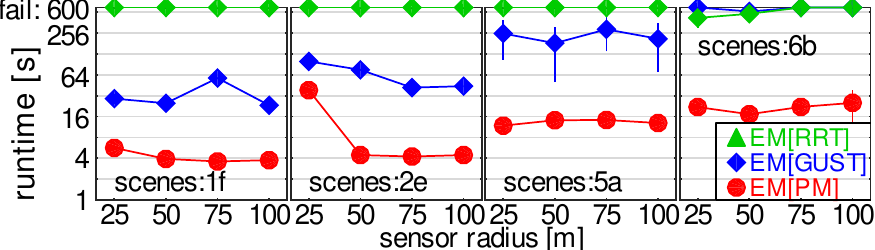}
    \includegraphics[width=0.49\textwidth]{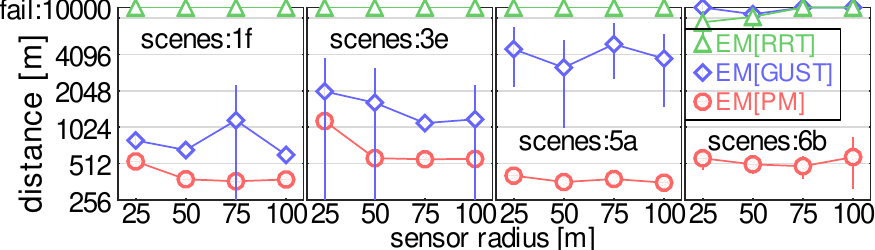}
    \caption{\textbf{Results when varying the sensor radius.}}
    \label{fig:ResSensor}
\end{figure*}

\subsection{Results}

\subsubsection{Runtime and Distance Results} Fig.~\ref{fig:ResTimeAndDist} summarizes the results for EM[PM], EM[GUST], and EM[RRT] for all six environment types and difficulty levels. EM[PM] is significantly faster than EM[GUST] and EM[RRT],  and capable of solving even the most challenging instances. In contrast, EM[RRT] struggles due to reliance on a nearest-neighbor heuristic, which often leads the explorations towards obstacles, particularly in obstacle-rich environments with narrow passages. EM[GUST] performs better than EM[RRT] as GUST is a more computationally efficient planner.

The distance results exhibit similar patterns to the runtime results. EM[PM] generates much shorter solutions compared to EM[GUST] and EM[RRT] since it is guided by the high-level discrete layer and strives for consistency between invocations. EM[GUST], despite also being guided by a high-level layer, displays more oscillatory behavior that leads to the robot moving back and forth between areas. EM[RRT] performs the worst as it either gets stuck or takes unnecessarily long routes to reach the goal.

\subsubsection{Impact of Leveraging Prior Solutions} Fig.~\ref{fig:ResNoHints} shows the results obtained when the prior solutions are disabled as inputs to the planning module. The resulting versions of the methods are referred to as EM[PM:no hints], EM[GUST:no hints], EM[RRT:no hints]. In this case, EM[PM:no hints] still performs significantly faster and produces shorter solutions than the other two methods. However, as expected, EM[PM:no hints] performs worse than EM[PM] due to the absence of prior solutions as a hint. This highlights the importance of incorporating prior solutions as hints, as it enables the planning module to reuse valid parts of the trajectory, resulting in a reduction of oscillatory behaviors and of overall runtime required to generate a new collision-free and dynamically-feasible trajectory to reach the goal.

\subsubsection{Results when Varying the Number of Snake Links} Fig.~\ref{fig:ResLinks} shows the results when varying the number of snake links. The results highlight the capabilities of EM[PM] to solve challenging problem instances. The performance for the other methods, EM[GUST] and EM[RRT], becomes considerably worse as the number of the snake links is increased, while EM[PM] remains efficient.

\subsubsection{Results when Varying the Sensor Range} Fig.~\ref{fig:ResSensor} shows the results when varying the sensor range. When the range is small, the execution becomes more challenging since the planning module is invoked more frequently to handle newly discovered obstacles. Moreover, the planned solutions tend to be closer to such obstacles. As the range increases, the robot can collect more information, leading to better plans. The results show that EM[PM] outperforms the other methods. By generating high-clearance solutions, EM[PM] requires fewer replanning calls, improving the overall runtime, and reducing the distance traveled by the robot.



\section{Discussion}
\label{sec:discussion}

This paper developed a framework for sampling-based dynamically-feasible motion planning well-suited for navigation in partially-known environments.
We have shown that our approach, which increases clearance to known obstacles and reuses prior solutions to guide motion-tree exploration when invoked to plan from a new state, significantly improves both runtime and solution quality across hundreds of structured and unstructured virtual environments with varying levels of difficulty and size.
In future work, we will use machine learning to imbue our planner with the ability to estimate occupancy information or large-scale structure of unseen space, information that it may use to further improve the robot's plans despite uncertainty about the environment.
\deleted{Other possible directions may involve extending the framework to support concurrent planning for multiple agents or to reach multiple goals in unseen space.}

\bibliography{refs,refs_gjs}
\bibliographystyle{IEEEtran}

\end{document}